\documentclass[12pt,a4paper]{article}
\usepackage[margin=1in]{geometry}
\usepackage{amsmath}
\usepackage{amssymb,amsfonts}
\usepackage{graphicx}
\usepackage{booktabs}
\usepackage{array}
\usepackage{setspace}
\usepackage{lineno}
\usepackage[hidelinks]{hyperref}
\usepackage{url}

\begin{document}

\begin{center}
  {\Large\bfseries Strip Convolution and Direction-Aware Exclusion Loss for Oriented Ship Detection}

  \vspace{1em}

  Bin Chen\textsuperscript{1,2}, Yuanyuan Liu\textsuperscript{1,*}, Peng Yang\textsuperscript{2}, Chao Lu\textsuperscript{2}

  \vspace{0.5em}

  {\small
  \textsuperscript{1}School of Information and Software Engineering, East China Jiaotong University, Nanchang 330013, China\\
  \textsuperscript{2}Jiangxi Vocational University of Foreign Studies, Nanchang 330099, China\\[0.3em]
  *Corresponding author. Email: lyy.78@163.com (Y. Liu). E-mail: chenbin\_ckf@163.com (B. Chen).
  }
\end{center}

\vspace{1em}

\begin{abstract}
Oriented ship detection in very high resolution (VHR) remote sensing imagery remains challenging due to elongated hull geometry and dense target distributions in complex port scenes. Existing methods typically address geometric representation and duplicate suppression separately. To jointly tackle these issues, we propose an oriented ship detector with two complementary components. The C3k2\_Strip module employs orthogonal strip convolutions to better capture elongated hull structures, while the Class-Aware Direction-Aware Exclusion Loss (CA-DAEL) suppresses redundant predictions using class, direction, and confidence cues. Experiments on HRSC2016 and DIOR-R achieve 78.45\% and 53.71\% mAP$_{\mathrm{50:95}}$, respectively, with only 2.91M parameters. On HRSC2016, the proposed method improves mAP$_{\mathrm{50:95}}$ by 6.32 percentage points over the YOLOv11-OBB baseline, demonstrating its effectiveness for accurate oriented ship detection.
\end{abstract}

% IJRS style: no keywords.

% =================================================================
\section{Introduction}
\label{sec:introduction}

\begin{figure}[!t]
  \centering
  \includegraphics[width=0.85\linewidth,trim={9 84 8 145},clip]{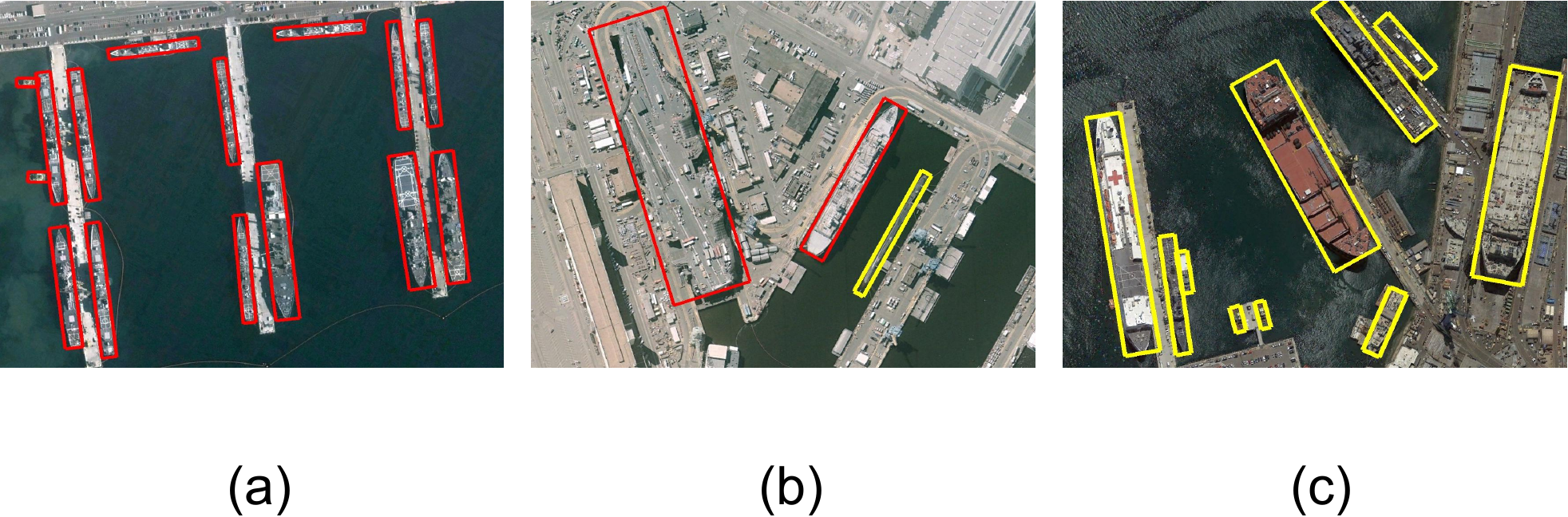}
  \caption{Challenging scenarios in oriented ship detection (ground truth in red). (a) Dense dockside overlap. (b) Slender vessel with aspect ratio $>$10:1. (c) OBB overlap between adjacent vessels.}
  \label{fig:challenge}
\end{figure}

Oriented ship detection in very high resolution (VHR) remote sensing imagery is important for maritime monitoring and security~\cite{sun2025fine,huo2025chgaff,huang2025dmsda}. Despite recent advances~\cite{han2021align,yang2021r3det,zhou2022mmrotate}, two challenges remain prominent, as shown in Fig.~\ref{fig:challenge}. First, elongated hulls are poorly matched to isotropic receptive fields, weakening directional representations. Second, densely docked ships cause severe OBB overlap, complicating regression and duplicate suppression. These coupled challenges limit accurate oriented ship detection.

The first challenge lies in geometric representation. Ship hulls can exhibit aspect ratios of up to 15:1, whereas standard $3\times3$ convolutions expand receptive fields uniformly in all directions, allocating substantial capacity to regions unrelated to the hull principal axis. This mismatch weakens directional responses for long and narrow vessels. Deformable convolution~\cite{dai2017deformable} adapts sampling locations to target geometry, while snake convolution~\cite{qi2023dynamic} provides flexible directional modeling, but both introduce additional computational or sequential overhead. ACNet~\cite{ding2019acnet} strengthens directional responses through asymmetric kernels, yet is not specifically designed for oriented ship detection. Strip-shaped operators have proven effective in broader contexts: Strip Pooling~\cite{hou2020strip} captures long-range context for scene parsing, and LSKNet~\cite{li2023lsk} adapts large selective kernels to remote-sensing backbones. However, both deploy strip or large kernels as standalone pooling or backbone components; in contrast, we embed deterministic orthogonal strip convolutions as a lightweight attention prior inside C3k2 bottlenecks and, crucially, couple the resulting direction-aligned features with a direction-aware exclusion loss so that geometric representation and duplicate suppression reinforce each other during training. Recent ship detectors also report feature degradation for elongated vessels~\cite{kim2025lim,zhu2025efficient,zhang2023ofcos}. These observations motivate an efficient operator that better aligns receptive fields with elongated hull geometry.

The second challenge arises from dense ship distributions in crowded ports, where substantial OBB overlap between neighboring vessels increases localization ambiguity and often leads to multiple redundant predictions on the same hull.
Such redundancy can bias box regression and reduce the reliability of non-maximum suppression (NMS). Two-stage methods such as RoI Transformer~\cite{ding2019learning} and Oriented R-CNN~\cite{xie2021oriented}, as well as single-stage detectors such as R3Det~\cite{yang2021r3det} and S2ANet~\cite{han2021align}, improve oriented localization but typically rely on relatively heavy architectures. Compact alternatives such as RTMDet-R~\cite{lyu2022rtmdet} reduce model complexity, yet duplicate predictions remain problematic in dense scenes.
From the optimization perspective, Repulsion Loss~\cite{wang2018repulsion} suppresses predictions near neighboring objects but is designed for horizontal boxes, while recent crowd-aware objectives do not jointly consider orientation, class identity, and prediction confidence. Therefore, an OBB-specific exclusion mechanism is needed to suppress redundant predictions without mistakenly repelling legitimate adjacent ships.

To address these two challenges jointly, we develop an oriented ship detector based on YOLOv11-OBB~\cite{khanam2024yolov11} with two complementary components. The C3k2\_Strip module augments standard bottlenecks with orthogonal strip convolutions, extending receptive fields along horizontal and vertical directions to better capture elongated hull structures. The Class-Aware Direction-Aware Exclusion Loss (CA-DAEL) further incorporates class, orientation, overlap, and confidence cues to suppress redundant same-class predictions while preserving valid neighboring vessels. The two components operate complementarily: C3k2\_Strip improves orientation-aligned feature representation, while CA-DAEL exploits more reliable directional cues to reduce prediction redundancy during training.
The main contributions are summarized as follows:
\begin{itemize}
    \item To alleviate the geometric mismatch between isotropic receptive fields and elongated hull structures, we introduce C3k2\_Strip, which employs orthogonal strip convolutions to strengthen direction-aligned feature representation with minimal computational overhead.

    \item To suppress redundant predictions in crowded port scenes, we develop CA-DAEL, a loss function that jointly considers class identity, orientation, overlap, and confidence while preserving legitimate adjacent ships.

    \item Extensive experiments on HRSC2016 and DIOR-R demonstrate the effectiveness and generalization of the proposed method, achieving 78.45\% and 53.71\% mAP$_{\mathrm{50:95}}$, respectively, with only 2.91M parameters.
\end{itemize}

% =================================================================
\section{Proposed Method}
\label{sec:method}
% =================================================================

\begin{figure}[!t]
  \centering
  \includegraphics[width=0.95\linewidth,trim={43 38 71 0},clip]{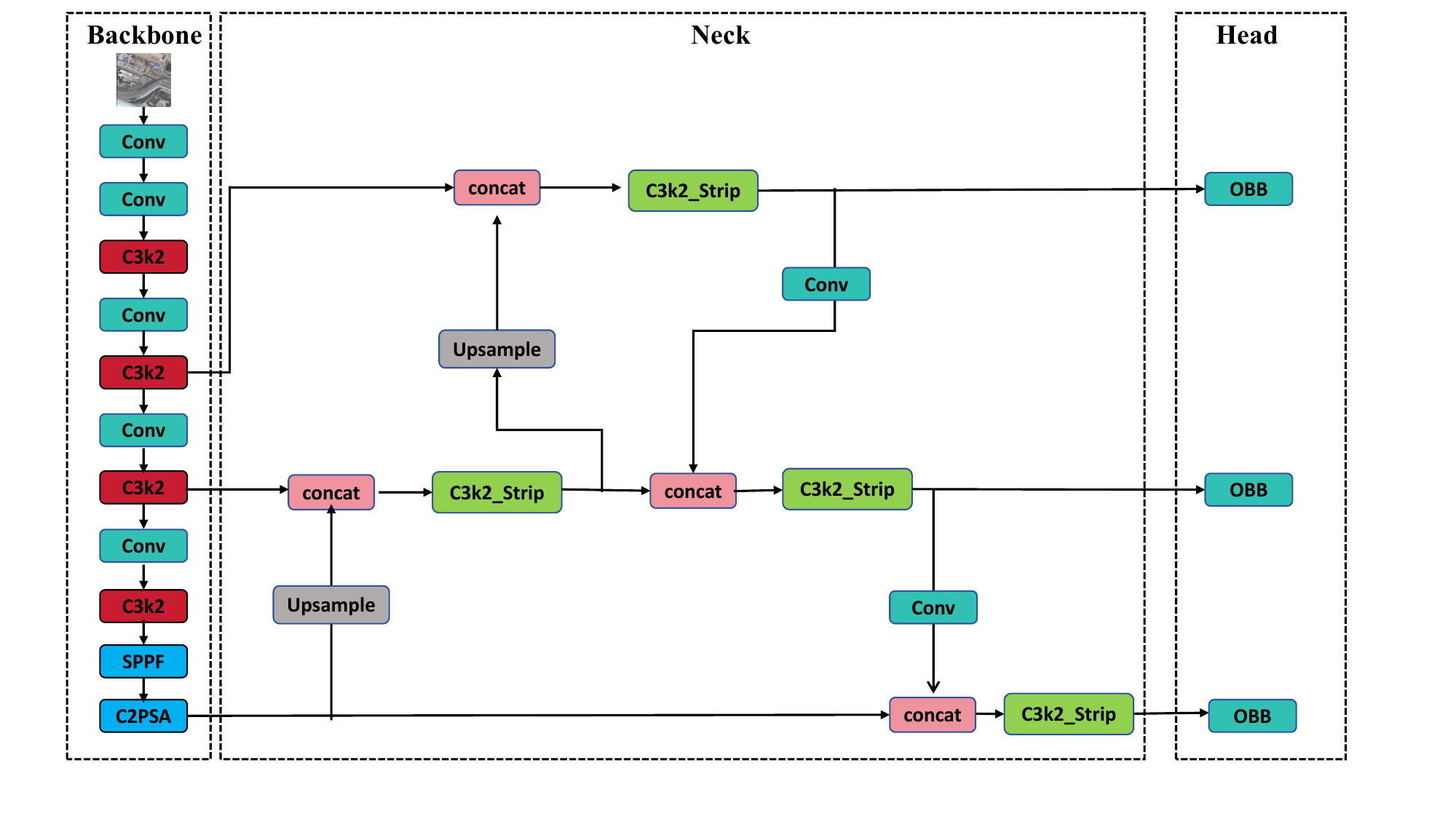}
  \caption{Overall architecture. A CSPDarknet backbone extracts multi-scale features P3--P5; the neck employs C3k2\_Strip modules that augment standard bottlenecks with orthogonal strip kernels. Three OBB heads output rotated boxes. The CA-DAEL loss is training-only.}
  \label{fig:framework}
\end{figure}

\noindent\textbf{Overall Framework.}
The end-to-end pipeline (Fig.~\ref{fig:framework}) follows the rotated-YOLO single-stage anchor-free paradigm. The CSPDarknet backbone, rooted in the YOLO-series design \cite{bochkovskiy2020yolov4}, and the OBB heads are inherited from YOLOv11-OBB (baseline); we introduce two \textit{novel components}: \textbf{C3k2\_Strip} modules (Sec.~\ref{sub:strip}, $k_2=7$) and the \textbf{CA-DAEL} loss (Sec.~\ref{sub:loss}). The heads output $(x, y, w, h, \theta)$ boxes, with $\theta\in[0^\circ, 180^\circ)$ under the long-edge convention. The model comprises 2.91M parameters and 7.25 GFLOPs.

\subsection{C3k2\_Strip Module for Elongated Features}
\label{sub:strip}
\begin{figure}[!t]
  \centering
  \includegraphics[width=0.95\linewidth,trim={12 112 29 140},clip]{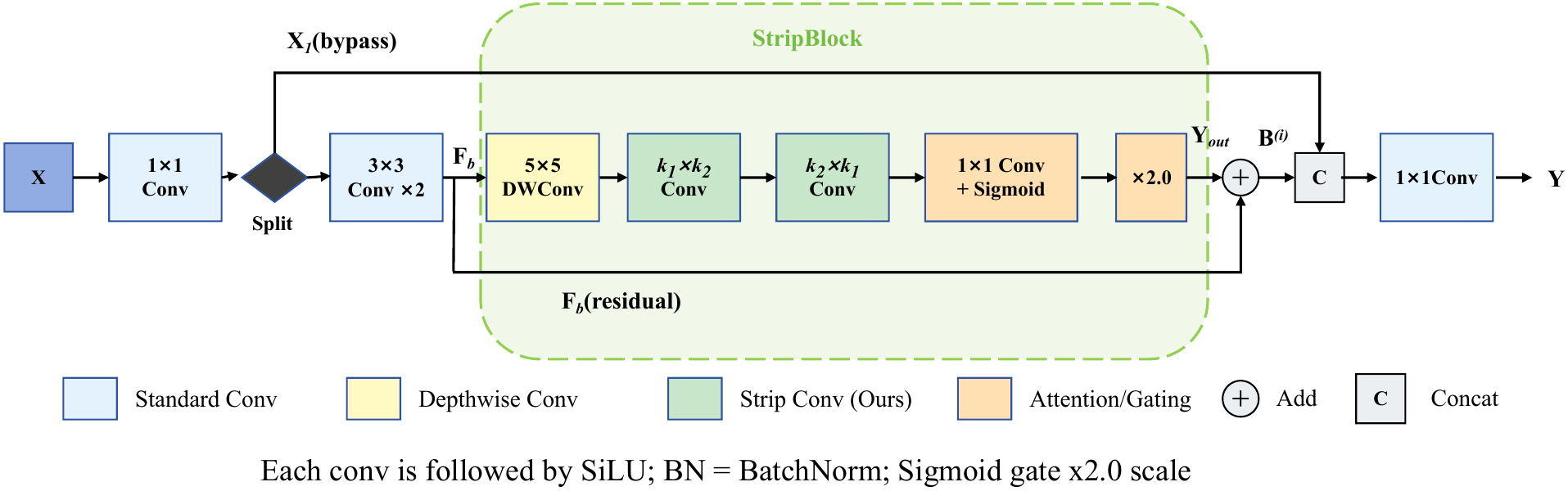}
  \caption{Structure of C3k2\_Strip. A $1\times1$ convolution splits the input into $\mathbf{X}_1$ (bypass) and $\mathbf{X}_2$ (processed). The Bottleneck\_Strip produces $\mathbf{F}_b$ via two $3\times3$ convolutions (each followed by SiLU), then the StripBlock (highlighted) computes attention via $5\times5$ DWConv $\to$ orthogonal $k_1\times k_2$ and $k_2\times k_1$ strip convs $\to$ $1\times1$ Conv + BN + Sigmoid, modulated by a $2.0$ scale. The residual sum $\mathbf{F}_b + \mathbf{Y}_{\text{att}}$ forms $\mathbf{B}^{(i)}$; all branches are fused by a final $1\times1$ convolution.}
  \label{fig:strip}
\end{figure}

A standard bottleneck stacks two $3\times 3$ convolutions whose isotropic receptive fields waste capacity on regions orthogonal to the hull principal axis when targets are elongated. Rather than discarding these kernels, we \emph{augment} each bottleneck with a \textit{strip-convolution attention block} (Fig.~\ref{fig:strip}) that selectively enhances hull-aligned responses, keeping the $3\times3$ convolutions for general feature transformation and letting the strip block act as a complementary attention prior. Let $\mathbf{F}_b$ denote the bottleneck output (per-unit superscripts dropped for brevity). A $5\times5$ depthwise preprocessing step first expands the local context along both axes:

\begin{equation}
  \mathbf{F}_0 = \text{SiLU}\!\left(\text{DWConv}_{5\times 5}(\mathbf{F}_b)\right),
  \label{eq:strip_pre}
\end{equation}

\noindent where $\text{DWConv}_{5\times 5}$ denotes depthwise convolution and $\text{SiLU}$ is the Sigmoid Linear Unit activation. Two orthogonal strip convolutions with kernels $k_1\times k_2$ (horizontal) and $k_2\times k_1$ (vertical) are then applied sequentially to capture hull-aligned features along both principal axes:

\begin{equation}
  \mathbf{F}_{\text{strip}} = \text{SiLU}\!\left(\text{Conv}_{k_2\times k_1}\!\left(\text{SiLU}\!\left(\text{Conv}_{k_1\times k_2}(\mathbf{F}_0)\right)\right)\right),
  \label{eq:stripconv}
\end{equation}

\noindent where $k_1=3$ and $k_2=7$ are fixed across all neck stages. The strip features $\mathbf{F}_{\text{strip}}$ are then compressed into a channel attention map via $1\times 1$ convolution and batch normalization (BN), which modulates the \emph{original bottleneck feature} $\mathbf{F}_b$ with a $2.0$ scale factor ensuring feature enhancement rather than suppression:

\begin{equation}
  \mathbf{Y}_{\text{att}} = \mathbf{F}_b \odot \sigma\!\left(\text{BN}\!\left(\text{Conv}_{1\times 1}(\mathbf{F}_{\text{strip}})\right)\right) \times 2.0,
  \label{eq:attn}
\end{equation}

\noindent where $\sigma(\cdot)$ is the sigmoid and $\odot$ denotes channel-wise multiplication. This design lets strip-convolved features serve as an attention prior that selectively amplifies hull-aligned responses in $\mathbf{F}_b$. The cross-stage partial (CSP) path first maps $\mathbf{X}$ through a $1\times 1$ convolution into $2c$ channels and splits it into $\mathbf{X}_1, \mathbf{X}_2$ (each $c$ channels). $\mathbf{X}_1$ bypasses directly; $\mathbf{X}_2$ passes sequentially through $n$ Bottleneck\_Strip units ($n$: per-stage C3k2 depth). Each unit uses an expansion ratio $e=1.0$ (no channel reduction), so both $3\times3$ convolutions operate at full width: $\mathbf{F}_b^{(i)} = \text{Conv}_{3\times3}(\text{Conv}_{3\times3}(\cdot))$. Since input and output channels match ($c_1=c_2=c$), a residual connection fuses the convolutional feature with the strip attention (Eqs.~\ref{eq:strip_pre}--\ref{eq:attn}):
\begin{equation}
  \mathbf{B}^{(i)} = \mathbf{F}_b^{(i)} + \mathbf{Y}_{\text{att}}^{(i)}, \quad i = 1,\dots,n,
  \label{eq:bottleneck}
\end{equation}
\noindent where $\mathbf{B}^{(i)}$ is the output of the $i$-th bottleneck and $\mathbf{Y}_{\text{att}}^{(i)}$ is given by Eq.~\ref{eq:attn} evaluated on $\mathbf{F}_b^{(i)}$. The final output fuses all branches via $1\times 1$ convolution: $\mathbf{Y} = \text{Conv}_{1\times 1}([\mathbf{X}_1, \mathbf{X}_2, \mathbf{B}^{(1)}, \dots, \mathbf{B}^{(n)}])$, where $[\cdot]$ denotes channel concatenation.
The strip length $k_2=7$ is fixed across all neck stages, providing elongated $7\times3$ and $3\times7$ kernels aligned with ship principal axes. This value is matched to vessel scale: at neck strides 8, 16, and 32 (stages P3--P5), a $7$-pixel kernel spans 56, 112, and 224 input pixels, so a 300-m tanker ($\sim$200 pixels at 1.5-m ground sample distance (GSD)) is fully covered at P5 and smaller vessels at P4 and P3. Unlike Snake and Deformable convolutions \cite{dai2017deformable,qi2023dynamic}, strip kernels are deterministic and hardware-friendly, and the model comprises 2.91M parameters / 7.25 GFLOPs (baseline: 2.66M / 6.69 GFLOPs)---an 8.4\% FLOPs overhead for $+4.41$ mAP$_{\mathrm{50:95}}$ (Table~\ref{tab:ablation}, R1$\to$R2).

\subsection{Class-Aware Direction-Aware Exclusion Loss}
\label{sub:loss}
\begin{figure}[!t]
  \centering
  \includegraphics[width=0.8\linewidth,trim={2 5 5 4},clip]{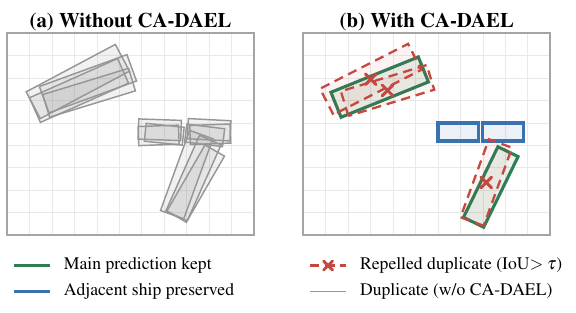}
  \caption{Mechanism of CA-DAEL on a dense dock scene. (a) Without CA-DAEL, the detector produces duplicate same-class predictions on the same hull. (b) With CA-DAEL, same-class pairs exceeding $\tau{=}0.8$ IoU are repelled (red dashed), the winning prediction on each hull is kept (green), and lower-overlap detections on distinct adjacent vessels are preserved (blue). Key parameters: $\tau{=}0.8$, $\lambda_{\text{exc}}{=}0.3$, 100-epoch loss-specific warmup (separate from the 3-epoch LR warmup).}
  \label{fig:cadael}
\end{figure}

Conventional detection losses treat every prediction independently and overlook the redundancy that arises when several overlapping boxes land on the same target---a deficiency that NMS cannot fully repair in dense port scenes. To close this gap, we introduce the \textit{Class-Aware Direction-Aware Exclusion Loss (CA-DAEL)}, which augments the standard rotated-YOLO losses with a class/direction/confidence-gated exclusion term:

\begin{equation}
  \mathcal{L}_{\text{CA-DAEL}} = \lambda_r \mathcal{L}_{\text{reg}} + \lambda_c \mathcal{L}_{\text{cls}} + \lambda_d \mathcal{L}_{\text{dfl}} + \lambda_a \mathcal{L}_{\text{angle}} + \lambda_{\text{exc}} \mathcal{L}_{\text{exc}},
  \label{eq:total_loss}
\end{equation}

\noindent where $\mathcal{L}_{\text{reg}}$ is the rotated complete intersection over union (CIoU) regression loss, $\mathcal{L}_{\text{cls}}$ is the binary cross-entropy (BCE) classification loss, $\mathcal{L}_{\text{dfl}}$ is the Distribution Focal Loss (DFL) for box edge refinement, and $\mathcal{L}_{\text{angle}}$ is the aspect-ratio-weighted angle loss inherited from the YOLOv11-OBB baseline:

\begin{equation}
  \mathcal{L}_{\text{angle}} = \frac{1}{N_{\text{fg}}}\sum_{i} \exp\!\left(-\frac{\ln^2(w_i^{*}/h_i^{*})}{\lambda_\theta^2}\right) \sin^2\!\big(2\,\widetilde{\Delta\theta}_i\big),
  \label{eq:angle}
\end{equation}

\noindent where $\widetilde{\Delta\theta}_i = (\hat{\theta}_i - \theta_i^{*}) - \text{round}((\hat{\theta}_i - \theta_i^{*})/\pi)\cdot\pi$ is the angle difference wrapped to $[-\pi/2, \pi/2]$, $w_i^{*}$ and $h_i^{*}$ are the ground-truth box dimensions, and $\lambda_\theta{=}3$ controls the aspect-ratio sensitivity. The $\sin^2(2\cdot)$ form peaks at $45^\circ$ (maximum ambiguity between long and short axes) and vanishes at $0^\circ$ and $90^\circ$. The aspect-ratio weight $\exp(-\ln^2(w^*/h^*)/\lambda_\theta^2)$ decreases for elongated objects, since their angles are already well-constrained by the CIoU regression loss---small angular errors cause large IoU drops for high-AR boxes---and $N_{\text{fg}}$ normalizes by the number of foreground targets. The loss weights follow the Ultralytics defaults ($\lambda_r{=}7.5$, $\lambda_c{=}0.5$, $\lambda_d{=}1.5$, $\lambda_a{=}1.0$). The exclusion term $\mathcal{L}_{\text{exc}}$ combines five gated factors---an IoU gate, overlap strength, a direction factor, a confidence factor, and a class gate:
\begin{equation}
\begin{aligned}
  \mathcal{L}_{\text{exc}} ={}& \frac{1}{|\mathcal{P}_{\text{act}}|}\sum_{(i,j)\in\mathcal{P}_{\text{act}}}
  \max(0, \text{IoU}_{ij} - \tau)^2 \cdot \text{IoU}_{ij} \\
  & \cdot (1 + \alpha_w \cos\widetilde{\Delta\theta}_{ij}) \cdot [s_i(1-s_j) + s_j(1-s_i)] \\
  & \cdot \mathbf{1}_{[c_i = c_j]},
\end{aligned}
  \label{eq:exclusion}
\end{equation}
\noindent where $\mathcal{P}_{\text{act}}$ is the set of same-class prediction pairs with non-zero penalty (i.e., pairs exceeding the IoU threshold $\tau{=}0.8$), $\text{IoU}_{ij}$ is the Gaussian-model-based probabilistic rotated IoU \cite{yang2021gwd}, computed as in the MMRotate implementation \cite{zhou2022mmrotate}, $\alpha_w{=}0.5$ is the angle weight, $\widetilde{\Delta\theta}_{ij}$ is the angular difference between predictions $i$ and $j$ normalized to $[0, \pi/2]$ via modulo-$\pi$ folding, $s_i, s_j$ are the confidence scores of predictions $i, j$, $c_i, c_j$ are their predicted class labels, $\mathbf{1}_{[c_i=c_j]}$ is the class-aware indicator, and $\lambda_{\text{exc}}=0.3$ is the overall loss weight. The exclusion term is activated after a 100-epoch loss-specific warmup to avoid early-training interference, distinct from the 3-epoch learning-rate warmup used for overall optimization.
The gating factors are designed around the structure of dense dock scenes. A high threshold $\tau=0.8$ avoids penalizing legitimate adjacent vessels, whose inter-OBB IoU typically stays in the 0.6--0.7 range; the direction factor strengthens penalties for parallel pairs; the confidence factor enforces one-directional suppression; and the class indicator prevents cross-class interference. Crucially, the direction factor is most informative when predicted angles $\hat{\theta}$ are well-aligned with the true hull axis, exactly what the strip features of Sec.~\ref{sub:strip} provide, which is the principal motivation for coupling CA-DAEL with C3k2\_Strip rather than using it in isolation. Fig.~\ref{fig:cadael} illustrates the resulting mechanism: same-class prediction pairs exceeding the IoU threshold are repelled, while lower-overlap or cross-class detections are preserved.

% =================================================================
\section{Experiments}
\label{sec:experiments}
% =================================================================

\subsection{Datasets and Evaluation Protocol}
\label{sub:datasets}
\textbf{HRSC2016} \cite{liu2017high}: 1,061 images, 2,976 ship instances (train/val/test: 436/181/444).
\textbf{DIOR-R} \cite{li2020object,cheng2022anchor}: 23,463 images, 192,472 instances in 20 categories (train/val/test: 11,725/1,178/10,560). Evaluation uses polygon-based rotated IoU following the MMRotate protocol \cite{zhou2022mmrotate}.

\subsection{Implementation Details}
\label{sub:impl}
Experiments were implemented using the Ultralytics framework with YOLOv11-OBB as the baseline. Hardware: Intel Xeon E5-2680 v4, 64 GB RAM, NVIDIA RTX 3080 Ti (12 GB); software: Python 3.10.20, PyTorch 2.5.1, CUDA 12.1. All models used $640\times640$ input, AdamW optimizer (LR 0.002, momentum 0.937, weight decay $5\times10^{-4}$), linear LR scheduling, batch size 32, 200 epochs on HRSC2016 and 300 on DIOR-R with a 3-epoch learning-rate warm-up. A separate 100-epoch CA-DAEL warm-up delays the exclusion term until regression and classification stabilize. Data augmentation includes extended rotation ($\pm30^\circ$), shear deformation ($\pm10^\circ$), and standard Mosaic~\cite{bochkovskiy2020yolov4} (disabled in the last 10 epochs). Loss weights are tuned on the HRSC2016 validation set.

\begin{table}[t]
  \centering
  \caption{Quantitative comparison with state-of-the-art methods on HRSC2016. All methods are trained for 200 epochs at $640\times640$ input resolution. All YOLO-series rotated detectors (YOLOv8~\cite{jocher2023yolov8}, YOLO26~\cite{jocher2026yolo26}, and YOLOv11~\cite{khanam2024yolov11}) are nano-scale variants. $\dagger$ denotes the proposed method. Best-performing values are highlighted in \textbf{bold}.}
  \label{tab:sota}
  \resizebox{0.85\linewidth}{!}{
  \begin{tabular}{lccccc}
    \toprule
    Method & Params(M) & P(\%) & R(\%) & mAP$_{\mathrm{50}}$ & mAP$_{\mathrm{50:95}}$ \\
    \midrule
    \multicolumn{6}{l}{\textit{Generic rotated detectors (non-YOLO)}} \\
    S2ANet \cite{han2021align}         & 41.12 & 90.53 & 85.59 & 86.70 & 44.79 \\
    R3Det \cite{yang2021r3det}          & 41.68 & 91.01 & 84.04 & 87.00 & 46.01 \\
    RTMDet-R-tiny \cite{lyu2022rtmdet} & \phantom{0}4.87 & 79.04 & 76.79 & 74.50 & 36.43 \\
    \midrule
    \multicolumn{6}{l}{\textit{YOLO-series rotated detectors}} \\
    YOLOv8-OBB & \phantom{0}3.08 & 90.39 & 84.26 & 91.98 & 74.08 \\
    YOLO26-OBB & \phantom{0}2.65 & 86.27 & 81.35 & 87.30 & 70.91 \\
    YOLOv11-OBB (baseline)            & \phantom{0}2.66 & 87.88 & 85.83 & 91.21 & 72.13 \\
    \midrule
    \textbf{Ours}$^\dagger$ & \phantom{0}2.91 & 89.11 & \textbf{87.92} & \textbf{94.67} & \textbf{78.45} \\
    \bottomrule
  \end{tabular}
  }
\end{table}

\begin{table}[t]
  \centering
  \caption{Cross-dataset generalization results on the DIOR-R benchmark, reported as the 20-class mean over all categories. All methods are trained for 300 epochs at $640\times640$ input resolution. All YOLO-series detectors are nano-scale variants. $\dagger$ denotes the proposed method. Best-performing values are highlighted in \textbf{bold}.}
  \label{tab:dior}
  \resizebox{0.7\linewidth}{!}{
  \begin{tabular}{lcccc}
    \toprule
    Method & P(\%) & R(\%) & mAP$_{\mathrm{50}}$ & mAP$_{\mathrm{50:95}}$ \\
    \midrule
    YOLOv8-OBB & 80.43 & 62.11 & 66.23 & 50.14 \\
    YOLO26-OBB & 81.22 & 62.66 & 67.10 & 51.21 \\
    YOLOv11-OBB (baseline)            & 81.28 & 62.13 & 66.47 & 50.34 \\
    \midrule
    \textbf{Ours}$^\dagger$ & \textbf{82.34} & \textbf{63.84} & \textbf{71.47} & \textbf{53.71} \\
    \bottomrule
  \end{tabular}
  }
\end{table}

\subsection{Comparison with State-of-the-Art Methods}
\label{sub:sota}
\begin{figure}[!t]
  \centering
  \includegraphics[width=0.8\linewidth]{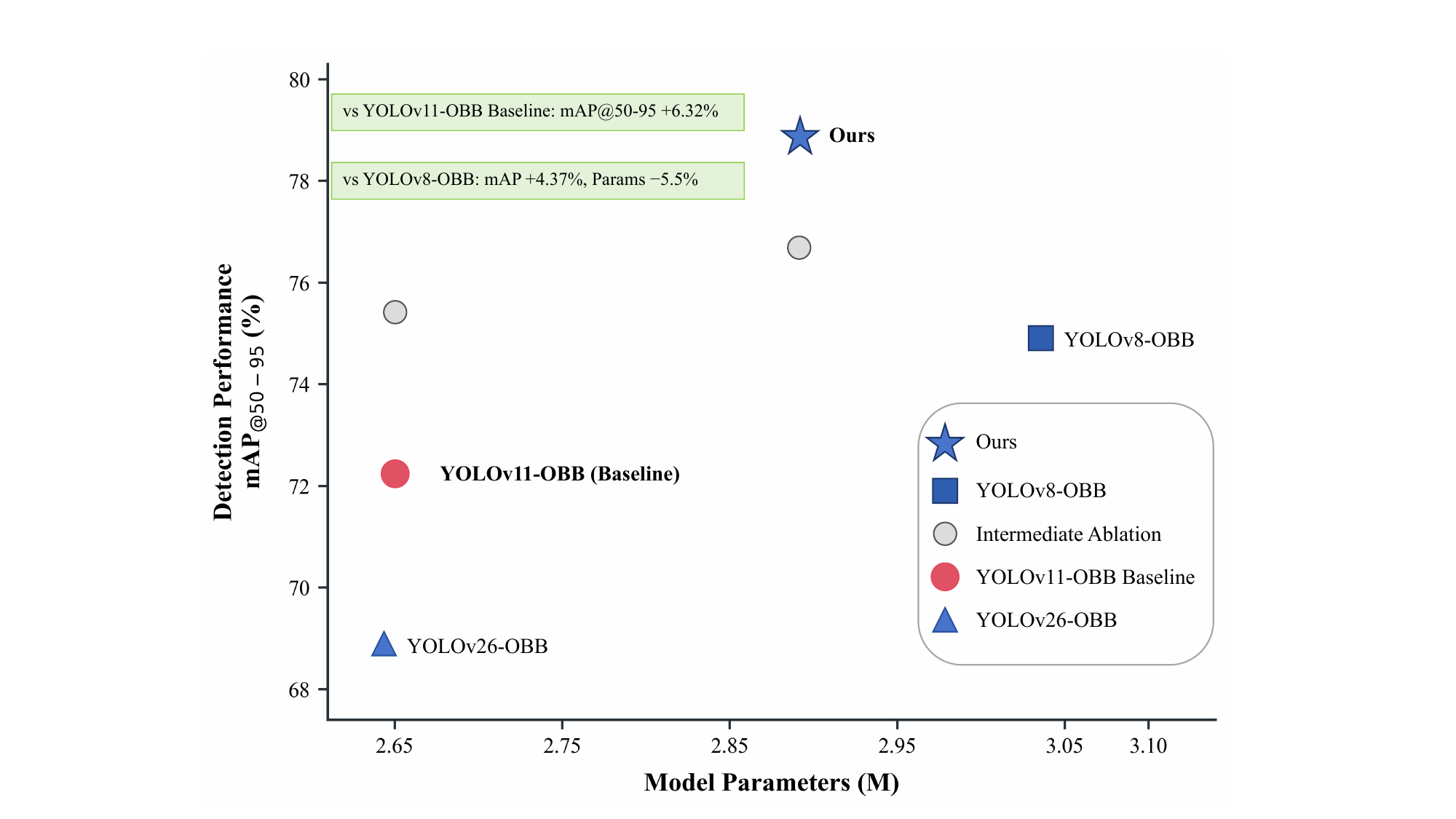}
  \caption{Parameter--accuracy comparison within the YOLO-series rotated detectors on HRSC2016. The proposed model attains the highest mAP$_{\mathrm{50:95}}$ under 3M parameters; gray points denote the two intermediate ablation configurations (R2, R3) in Table~\ref{tab:ablation}.}
  \label{fig:param_map}
\end{figure}

All methods in Tables~\ref{tab:sota}--\ref{tab:dior} were re-implemented and trained under identical protocols (200 epochs on HRSC2016, 300 epochs on DIOR-R, $640\times640$ input, AdamW optimizer, polygon-based mAP), which eliminates cross-paper protocol mismatches \cite{zhou2022mmrotate}.
On HRSC2016 (Table~\ref{tab:sota}), the proposed method reaches 78.45\% mAP$_{\mathrm{50:95}}$, outperforming YOLOv8-OBB by 4.37 points and leaving S2ANet and R3Det more than 32 points behind. Its recall of 87.92\%, the highest among all compared methods, indicates that CA-DAEL suppresses duplicates in dense dock scenes without sacrificing detection sensitivity, while strip-aligned features sustain competitive precision (89.11\%) on elongated hulls. As Fig.~\ref{fig:param_map} shows, the 2.91M-parameter / 7.25-GFLOP budget sits on the Pareto frontier of the YOLO family: no sub-3M model matches its accuracy. As Table~\ref{tab:sota} shows, generic detectors with $>$4M parameters still trail in mAP$_{\mathrm{50:95}}$.

\subsection{Cross-Dataset Generalization on DIOR-R}
\label{sub:dior}
To evaluate generalization beyond HRSC2016, we retrain on DIOR-R \cite{li2020object,cheng2022anchor}, a 20-category oriented benchmark with heterogeneous targets whose OBB annotations follow the anchor-free oriented proposal generator (AOPG) \cite{cheng2022anchor}, and report the 20-class mean.
Transferred to DIOR-R (Table~\ref{tab:dior}), the proposed method attains 71.47\% mAP$_{\mathrm{50}}$ and 53.71\% mAP$_{\mathrm{50:95}}$, improving over YOLOv11-OBB by 5.00 and 3.37 points. The gains are not ship-specific: C3k2\_Strip generalizes to heterogeneous elongated targets such as vehicles and aircraft, while CA-DAEL remains advantageous on categories with frequent spatial overlap. Both mechanisms therefore generalize across datasets, which suggests that elongated-feature alignment and dense-arrangement suppression capture structural rather than dataset-specific priors.

\subsection{Ablation Studies}
\label{sub:ablation}
Table~\ref{tab:ablation} isolates the contributions of the two proposed modules through four 200-epoch runs.
Adding C3k2\_Strip alone (R2) lifts mAP$_{\mathrm{50:95}}$ by $+4.41$ points at a cost of just 0.25M parameters and 0.56 GFLOPs, confirming that deterministic strip kernels effectively capture elongated hull geometry. CA-DAEL alone (R3) contributes $+3.65$ mAP$_{\mathrm{50:95}}$ with \emph{zero} parameter overhead, showing that the class and confidence gates already suppress most same-class duplicates without any inference-time cost.
Combining the two modules (R4) pushes the model to 78.45\% mAP$_{\mathrm{50:95}}$ ($+6.32$). The standalone gains sum to 8.06, yet the combined gain is 6.32, leaving 1.74 mAP$_{\mathrm{50:95}}$ of overlap ($3.65 - 1.91$). Rather than indicating redundancy, this overlap reflects that strip-aligned features already eliminate some of the duplicates that CA-DAEL targets---the two modules attack overlapping failure modes from complementary angles, which is precisely the synergy we set out to achieve.

\begin{table}[t]
  \centering
  \caption{Ablation study on HRSC2016 (200 epochs, $640{\times}640$). Strip denotes the C3k2\_Strip module; CA-DAEL denotes the exclusion loss. Params and GFLOPs measure model complexity.}
  \label{tab:ablation}
  \resizebox{0.8\linewidth}{!}{
  \begin{tabular}{@{}ccccccc@{}}
    \toprule
    ID & Strip & CA-DAEL & mAP$_{\mathrm{50}}$ & mAP$_{\mathrm{50:95}}$ & Params(M) & GFLOPs \\
    \midrule
    R1 & -- & -- & 91.21 & 72.13 & 2.66 & 6.69 \\
    R2 & \checkmark & -- & 93.13 & 76.54 & 2.91 & 7.25 \\
    R3 & -- & \checkmark & 92.70 & 75.78 & 2.66 & 6.69 \\
    R4 & \checkmark & \checkmark & \textbf{94.67} & \textbf{78.45} & 2.91 & 7.25 \\
    \bottomrule
  \end{tabular}
  }
\end{table}

\begin{figure}[!t]
  \centering
  \includegraphics[width=0.9\linewidth,trim={3 3 3 3},clip]{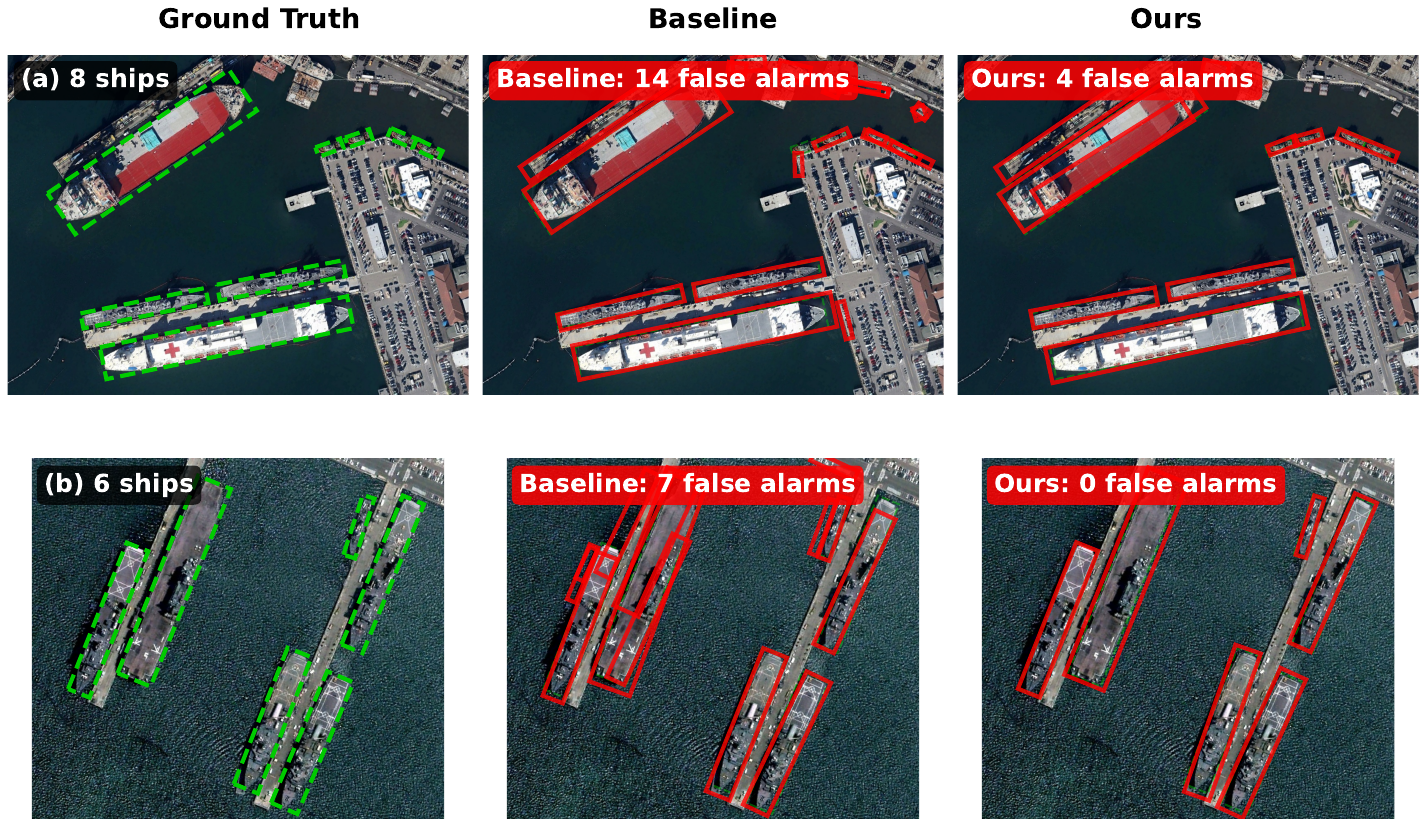}
  \caption{Qualitative comparison on HRSC2016 dense dock scenes (green dashed: ground truth; red solid: predictions). (a) An 8-ship scene. (b) A 6-ship scene with long-hull ships in tight parallel mooring.}
  \label{fig:qualitative}
\end{figure}

\subsection{Qualitative Analysis}
\label{sub:qualitative}
To verify that the two proposed modules address their intended challenges, we compare predictions on representative HRSC2016 test images featuring dense dock scenes with long-hull ships arranged in tight parallel strips (Fig.~\ref{fig:qualitative}).
The two scenes in Fig.~\ref{fig:qualitative} expose the failure modes that C3k2\_Strip and CA-DAEL respectively target. On the 8-ship scene (a), the baseline fragments each hull into multiple redundant boxes and produces OBBs whose angles deviate from the hull principal axis, a direct consequence of isotropic $3{\times}3$ kernels. The orthogonal $3{\times}7$ and $7{\times}3$ strip kernels realign the receptive field along that axis, yielding OBBs that hug the elongated geometry, while CA-DAEL's class/direction/confidence gating removes same-class duplicates, cutting false alarms from 14 to 4. The 6-ship scene (b) is even more telling: the baseline emits 7 false alarms, whereas the proposed method achieves zero false alarms with each ship correctly bounded. The qualitative behavior is thus consistent with the quantitative gains in Table~\ref{tab:ablation}: strip alignment fixes the angular drift, and exclusion gating cleans up the residual duplicates that alignment alone cannot remove.

% =================================================================
\section{Conclusion}
\label{sec:conclusion}
% =================================================================

This paper presented an oriented ship detector built on YOLOv11-OBB that tackles elongated hull geometry and dense dockside redundancy jointly. The C3k2\_Strip module augments standard bottlenecks with orthogonal strip convolutions that realign the receptive field along the hull principal axis, and the CA-DAEL loss exploits the resulting orientation-aligned features to repel same-class duplicate predictions through class, direction, and confidence gating at zero inference-time cost. The 2.91M-parameter model reaches 78.45\% mAP$_{\mathrm{50:95}}$ on HRSC2016 and 53.71\% on DIOR-R, surpassing YOLOv8-OBB by 4.37 points at only 7.25 GFLOPs. Future work will explore aspect-ratio-adaptive strip kernels that adjust kernel length to target elongation, together with multi-seed validation across diverse oriented object detection scenarios.

% =================================================================
% Required Taylor & Francis declaration sections
% TODO(投稿前确认): 如有基金资助，请在 Acknowledgements/Funding 中补充基金号
% =================================================================
\section*{Acknowledgements}
% 如有基金项目或需要致谢的个人/单位，请在此处填写；当前留空。

\section*{Disclosure statement}
The authors report there are no competing interests to declare.

\section*{Funding}
The authors received no financial support for the research, authorship, and/or publication of this article.

\section*{Data availability statement}
The datasets analysed in this study, HRSC2016 and DIOR-R, are publicly available benchmarks.

\section*{Author contributions}
Bin Chen: conceptualization, methodology, software, formal analysis, investigation, writing---original draft, and visualization. Yuanyuan Liu: supervision, writing---review and editing, and project administration. Peng Yang and Chao Lu: validation and writing---review and editing. All authors reviewed the results and approved the final version of the manuscript.

\bibliographystyle{IEEEtran}
\bibliography{IEEEabrv,ref}

\begin{thebibliography}{10}
\providecommand{\url}[1]{#1}
\csname url@samestyle\endcsname
\providecommand{\newblock}{\relax}
\providecommand{\bibinfo}[2]{#2}
\providecommand{\BIBentrySTDinterwordspacing}{\spaceskip=0pt\relax}
\providecommand{\BIBentryALTinterwordstretchfactor}{4}
\providecommand{\BIBentryALTinterwordspacing}{\spaceskip=\fontdimen2\font plus
\BIBentryALTinterwordstretchfactor\fontdimen3\font minus
  \fontdimen4\font\relax}
\providecommand{\BIBforeignlanguage}[2]{{%
\expandafter\ifx\csname l@#1\endcsname\relax
\typeout{** WARNING: IEEEtran.bst: No hyphenation pattern has been}%
\typeout{** loaded for the language `#1'. Using the pattern for}%
\typeout{** the default language instead.}%
\else
\language=\csname l@#1\endcsname
\fi
#2}}
\providecommand{\BIBdecl}{\relax}
\BIBdecl

\bibitem{sun2025fine}
Y.~Sun and S.~Li, ``Fine-grained object detection of satellite video in the
  frequency domain,'' \emph{IEEE Geoscience and Remote Sensing Letters},
  vol.~22, pp. 1--5, 2025.

\bibitem{huo2025chgaff}
X.~Huo, H.~Gao, B.~Huang, and G.~Chen, ``Chgaff-yolo: a cascade hybrid global
  adaptive feature fusion framework for real-time ocean internal wave
  detection,'' \emph{IEEE Geoscience and Remote Sensing Letters}, vol.~22, pp.
  1--5, 2025.

\bibitem{huang2025dmsda}
Z.~Huang, Z.~Xu, X.~Li, Y.~Zhang, Y.~Shi, Q.~Li, and H.~Fang, ``Dmsda-yolo:
  Dynamic multi-scale dilated attention for remote sensing object detection,''
  \emph{IEEE Geoscience and Remote Sensing Letters}, 2025.

\bibitem{han2021align}
J.~Han, J.~Ding, J.~Li, and G.-S. Xia, ``Align deep features for oriented
  object detection,'' \emph{IEEE transactions on geoscience and remote
  sensing}, vol.~60, pp. 1--11, 2021.

\bibitem{yang2021r3det}
X.~Yang, J.~Yan, Z.~Feng, and T.~He, ``R3det: Refined single-stage detector
  with feature refinement for rotating object,'' in \emph{Proceedings of the
  AAAI conference on artificial intelligence}, vol.~35, no.~4, 2021, pp.
  3163--3171.

\bibitem{zhou2022mmrotate}
Y.~Zhou, X.~Yang, G.~Zhang, J.~Wang, Y.~Liu, L.~Hou, X.~Jiang, X.~Liu, J.~Yan,
  C.~Lyu \emph{et~al.}, ``Mmrotate: A rotated object detection benchmark using
  pytorch,'' in \emph{Proceedings of the 30th ACM international conference on
  multimedia}, 2022, pp. 7331--7334.

\bibitem{dai2017deformable}
J.~Dai, H.~Qi, Y.~Xiong, Y.~Li, G.~Zhang, H.~Hu, and Y.~Wei, ``Deformable
  convolutional networks,'' in \emph{2017 IEEE international conference on
  computer vision (ICCV)}.\hskip 1em plus 0.5em minus 0.4em\relax IEEE, 2017,
  pp. 764--773.

\bibitem{qi2023dynamic}
Y.~Qi, Y.~He, X.~Qi, Y.~Zhang, and G.~Yang, ``Dynamic snake convolution based
  on topological geometric constraints for tubular structure segmentation,'' in
  \emph{2023 IEEE/CVF International Conference on Computer Vision
  (ICCV)}.\hskip 1em plus 0.5em minus 0.4em\relax IEEE, 2023, pp. 6047--6056.

\bibitem{ding2019acnet}
X.~Ding, Y.~Guo, G.~Ding, and J.~Han, ``Acnet: Strengthening the kernel
  skeletons for powerful cnn via asymmetric convolution blocks,'' in \emph{2019
  IEEE/CVF international conference on computer vision (ICCV)}.\hskip 1em plus
  0.5em minus 0.4em\relax IEEE, 2019, pp. 1911--1920.

\bibitem{hou2020strip}
Q.~Hou, L.~Zhang, M.-M. Cheng, and J.~Feng, ``Strip pooling: Rethinking spatial
  pooling for scene parsing,'' in \emph{2020 IEEE/CVF Conference on Computer
  Vision and Pattern Recognition (CVPR)}.\hskip 1em plus 0.5em minus
  0.4em\relax IEEE, 2020, pp. 4003--4012.

\bibitem{li2023lsk}
Y.~Li, Q.~Hou, Z.~Zheng, M.-M. Cheng, J.~Yang, and X.~Li, ``Large selective
  kernel network for remote sensing object detection,'' in \emph{2023 IEEE/CVF
  International Conference on Computer Vision (ICCV)}.\hskip 1em plus 0.5em
  minus 0.4em\relax IEEE, 2023.

\bibitem{kim2025lim}
S.-H. Kim, H.~Sim, Y.~Jung, O.-C. Jung, and Y.~Kim, ``{LiM-YOLO}: Less is more
  with pyramid level shift and normalized auxiliary branch for ship detection
  in optical remote sensing imagery,'' \emph{arXiv preprint arXiv:2512.09700},
  2025.

\bibitem{zhu2025efficient}
L.~Zhu, S.~Ren, B.~Lu, and Z.~Chen, ``An efficient and lightweight detector for
  complex backgrounds and multiscale objects in remote sensing images,''
  \emph{IEEE Geoscience and Remote Sensing Letters}, vol.~22, pp. 1--5, 2025.

\bibitem{zhang2023ofcos}
D.~Zhang, C.~Wang, and Q.~Fu, ``Ofcos: An oriented anchor-free detector for
  ship detection in remote sensing images,'' \emph{IEEE Geoscience and Remote
  Sensing Letters}, vol.~20, pp. 1--5, 2023.

\bibitem{ding2019learning}
J.~Ding, N.~Xue, Y.~Long, G.-S. Xia, and Q.~Lu, ``Learning roi transformer for
  oriented object detection in aerial images,'' in \emph{2019 IEEE/CVF
  Conference on Computer Vision and Pattern Recognition (CVPR)}.\hskip 1em plus
  0.5em minus 0.4em\relax IEEE, 2019, pp. 2844--2853.

\bibitem{xie2021oriented}
X.~Xie, G.~Cheng, J.~Wang, X.~Yao, and J.~Han, ``Oriented r-cnn for object
  detection,'' in \emph{2021 IEEE/CVF International Conference on Computer
  Vision (ICCV)}.\hskip 1em plus 0.5em minus 0.4em\relax IEEE, 2021, pp.
  3500--3509.

\bibitem{lyu2022rtmdet}
C.~Lyu, W.~Zhang, H.~Huang, Y.~Zhou, Y.~Wang, Y.~Liu, S.~Zhang, and K.~Chen,
  ``{RTMDet}: An empirical study of designing real-time object detectors,''
  \emph{arXiv preprint arXiv:2212.07784}, 2022.

\bibitem{wang2018repulsion}
X.~Wang, T.~Xiao, Y.~Jiang, S.~Shao, J.~Sun, and C.~Shen, ``Repulsion loss:
  Detecting pedestrians in a crowd,'' in \emph{2018 IEEE/CVF conference on
  computer vision and pattern recognition}.\hskip 1em plus 0.5em minus
  0.4em\relax IEEE, 2018, pp. 7774--7783.

\bibitem{khanam2024yolov11}
R.~Khanam and M.~Hussain, ``{YOLOv11}: An overview of the key architectural
  enhancements,'' \emph{arXiv preprint arXiv:2410.17725}, 2024.

\bibitem{bochkovskiy2020yolov4}
A.~Bochkovskiy, C.-Y. Wang, and H.-Y.~M. Liao, ``Yolov4: Optimal speed and
  accuracy of object detection,'' \emph{arXiv preprint arXiv:2004.10934}, 2020.

\bibitem{yang2021gwd}
X.~Yang, J.~Yan, Q.~Ming, W.~Wang, X.~Zhang, and Q.~Tian, ``Rethinking rotated
  object detection with gaussian wasserstein distance loss,'' in
  \emph{International conference on machine learning}.\hskip 1em plus 0.5em
  minus 0.4em\relax PMLR, 2021, pp. 11\,830--11\,841.

\bibitem{liu2017high}
Z.~Liu, L.~Yuan, L.~Weng, and Y.~Yang, ``A high resolution optical satellite
  image dataset for ship recognition and some new baselines,'' in
  \emph{International conference on pattern recognition applications and
  methods}, vol.~2.\hskip 1em plus 0.5em minus 0.4em\relax SciTePress, 2017,
  pp. 324--331.

\bibitem{li2020object}
K.~Li, G.~Wan, G.~Cheng, L.~Meng, and J.~Han, ``Object detection in optical
  remote sensing images: A survey and a new benchmark,'' \emph{ISPRS journal of
  photogrammetry and remote sensing}, vol. 159, pp. 296--307, 2020.

\bibitem{cheng2022anchor}
G.~Cheng, J.~Wang, K.~Li, X.~Xie, C.~Lang, Y.~Yao, and J.~Han, ``Anchor-free
  oriented proposal generator for object detection,'' \emph{IEEE Transactions
  on Geoscience and Remote Sensing}, vol.~60, pp. 1--11, 2022.

\bibitem{jocher2023yolov8}
G.~Jocher, A.~Chaurasia, and J.~Qiu, ``Ultralytics {YOLOv8},''
  \url{https://github.com/ultralytics/ultralytics}, 2023.

\bibitem{jocher2026yolo26}
G.~Jocher, J.~Qiu, M.~Liu, S.~Lyu, F.~C. Akyon, and M.~E. Kalfaoglu,
  ``Ultralytics {YOLO26}: Unified real-time end-to-end vision models,''
  \emph{arXiv preprint arXiv:2606.03748}, 2026.

\end{thebibliography}

\end{document}